\documentclass[10pt,twocolumn,letterpaper]{article}

\usepackage{cvpr}
\usepackage{times}
\usepackage{epsfig}
\usepackage{graphicx}
\usepackage{amsmath}
\usepackage{amssymb}
\usepackage[normalem]{ulem}

\usepackage[breaklinks=true,bookmarks=false]{hyperref}
\usepackage{float}
\cvprfinalcopy 

\def\cvprPaperID{6} 

\ifcvprfinal\fi
\begin{document}

\title{Using Player's Body-Orientation to Model Pass Feasibility in Soccer}

\author{A. Arbu{\'e}s-Sang{\"u}esa$^{1}$, A. Mart{\'\i}n$^{1}$,  J. Fern{\'a}ndez$^{2}$, C. Ballester$^{1}$, G. Haro$^{1}$\\
$^{1}$Universitat Pompeu Fabra, $^{2}$ Futbol Club Barcelona\\
{\tt\small adria.arbues@upf.edu}
}

\maketitle

\begin{abstract}
Given a monocular video of a soccer match, this paper presents a computational model to estimate the most feasible pass at any given time. The method leverages offensive player's orientation (plus their location) and opponents' spatial configuration to compute the feasibility of pass events within players of the same team. Orientation data is gathered from body pose estimations that are properly projected onto the 2D game field; moreover, a geometrical solution is provided, through the definition of a feasibility measure, to determine which players are better oriented towards each other. Once analyzed more than 6000 pass events, results show that, by including orientation as a feasibility measure, a robust computational model can be built, reaching more than 0.7 Top-3 accuracy. Finally, the combination of the orientation feasibility measure with the recently introduced Expected Possession Value metric is studied; promising results are obtained, thus showing that existing models can be refined by using orientation as a key feature. These models could help both coaches and analysts to have a better understanding of the game and to improve the players' decision-making process.
\end{abstract}

\section{Introduction}
Pep Guardiola, current Manchester City's soccer coach and former Futbol Club Barcelona's, said once that elder people claim that in yesteryear soccer you had to control the ball, then look and turn around, and finally, make the pass, while in today's faster version of soccer, players need first to look (and orient correctly) before controlling and passing the ball. Therefore, getting orientation metrics may help coaches to boost the performance of a team by designing optimal tactics according to players’ strengths and weaknesses. However, the concept of orientation is a complex concept without an exact definition, and during a soccer game, there are a total of/up to 22 players oriented in their own way at any given time during 90 minutes. In order to avoid the so-called concept of \textit{paralysis by analysis}, in this paper soccer events are filtered, hence including just pass events, which are the ones in where orientation takes the most important role according to Guardiola's words. The main contribution of this research is a computational model that, for each pass event, outputs the feasibility of receiving the ball for each potential candidate of the offensive team. The proposed model combines three different types of feasibility measures, defined on the grounding assumption that, among all potential receivers, the passer will move the ball to the (a) best oriented, (b) less defended and (c) closest available player. Orientation is obtained through a Computer Vision state-of-the art method \cite{arbuesOrientation}, which outputs an orientation value for each player by projecting the upper-torso pose parts in a 2D field. On top of these data, a novel feasibility measure is introduced to describe how good/bad the orientation fit between a passer and a potential receiver is. Given the location of all defenders, another feasibility metric is defined to establish how tough it is for the passer to move the ball to a particular player; this metric takes into account the distance of all defenders with respect to the passing line, which is defined by the relative angle in the 2D field that joins the passer and the receiver. Finally, pairwise distances among offensive players are used to construct a third feasibility measure based on the separation between players, hence assuming that players close to the ball have higher chances of receiving it than farther ones.\\ Results, expressed with Top-1 and Top-3 accuracy, show that the combination of all feasibility measures outperforms any of their individual performances, and that the model strongly benefits from the inclusion of the orientation feasibility measure. Moreover, existing state-of-the-art (SoA) models have been tested and compared, both before and after adding orientation as a feature to predict the outcome of passes, obtaining promising results which show that models can be confidently refined by adding these type of data. \\ The rest of the paper is organized as follows: in Section \ref{sec:SoA}, the related research is analyzed, including the details of the methods this research stems from; the proposed computational model is described in Section \ref{sec:ProMet}, along with all technical details. Feasibility results, discussion and possible combinations are studied in Section \ref{sec:Res}, and finally, conclusions are drawn in Section \ref{sec:Conc}. 



\section{Related work} \label{sec:SoA}

Since the irruption of Moneyball \cite{lewis2004moneyball}, sports clubs started conducting research about applied data science with the main purpose of boosting team performance. More concretely, the inclusion of tracking data proved to be crucial for the design of team strategies, so computer vision became (and still is) a hot topic in this research field. Lately, many contributions have been made towards geometric and semantic sports analysis \cite{maksai2016players,bertasius2017baller,thomas2017computer,felsen2017will,shih2017survey,senocak2018part,wu2019learning,chen2019sports,dwibedi2019temporal,cioppa2019arthus,stein2019movement,ran2019robust}, mostly driven by direct applications that might be useful for coaches in order to prepare optimal tactics. In particular, recent contributions in soccer such as \cite{rematas2018soccer,cioppa2019arthus,chen2019sports,fernandez2019decomposing,chawla2017classification} managed to better explain this sport analytically through tracking data, among others. However, authors claim that there is still a lack of contextualization due to undefined variables, such as player body orientation. 
To the best of our knowledge, the only method that aims to extract player body orientation over soccer video footage was published by Arbués-Sangüesa \textit{et al.}~\cite{arbuesOrientation}. This method computes players' orientation by combining: (a) the angle of the player with respect to the ball, with (b) an estimation of the body orientation as a 2D projection of the normal vector to the upper-torso. In order to do so, this work first uses OpenPose \cite{ramakrishna2014pose, wei2016convolutional, cao2017realtime} over the soccer video footage to detect player's body keypoints. Moreover, a Support Vector Machine model (based on color and geometrical feature vectors) is applied in order to ensure that OpenPose parts are not swapped.
This method achieves a median absolute error of 26 degrees/player, and three different types of orientation visualization tools are introduced: OrientSonars, Reaction and On-Field maps. In the presented article, this method is used to obtain the estimation of the orientation of each player on the 2D field. 

Moreover, soccer analysts have been struggling for many years to find a way to assign some value to the individual actions performed by each player, thus  obtaining specific metrics for each move. Different passing probability models and the quantification of concepts such as the pass risk/reward are introduced in \cite{gyarmati2016qpass, link2016real, power2017not}, and deep analysis of passing strategies are studied in \cite{gyarmati2015automatic,szczepanski2016beyond,chawla2017classification}; more concretely, \cite{hubacek} proposes a passing prediction model based on an end-to-end CNN approach. Note that none of the previous models take orientation into account. 
Furthermore, given that the main reward of soccer players is to score a goal, and knowing that this type of action is a rare event during the 90 minutes of the game, Fernandez \textit{et al.}  \cite{fernandez2019decomposing} introduced a new metric called Expected Possession Value (EPV), which already existed for basketball scenarios \cite{cervone2014pointwise}. The main objective of this metric is to predict an expected value of scoring/receiving a goal at a given time in any field position, based on a spatial analysis of the whole offensive and defensive setup at that moment; more concretely, in pass events, having a passer $P$, an EPV map can be computed for each field position $x\in\mathbb{R}^2$, which estimates the above-mentioned expected-value if $P$ passes the ball to $x$. The main EPV model consists of different likelihood components, especially emphasizing a passing probability model. In the present paper we will include a comparison and an analysis illustrating that those previous proposals can be improved by introducing player orientation information in the pass event analysis.

\section{Proposed Pass-Orientation Model} \label{sec:ProMet}
In this section, we propose a computational model to estimate the most plausible ball player pass at any given time based on the prior information that a player is going to execute a pass. To achieve this goal, we will attribute a feasibility score obtained by defining appropriate estimations that take into account player orientation 
and the configuration of the offensive and defensive team in the 2D field at that time. Intuitively, it stems from the fact that, in a pass event, 
there are 10 potential candidates of the same team who might receive the ball, each one of them holding a particular orientation with respect to the passer and at a certain position in the field. 

Let $u(\cdot,t)$ be a color video defined on $\Omega\times\left\{1,\dots, T\right\}$, where $\Omega\subset\mathbb{R}^2$ denotes the image frame domain and $\left\{1,\dots, T\right\}$ is the set of discrete times.
Given a time $t$, our method first considers the visible players in $u(\cdot,t)$ (\textit{i.e.}, visible players in the image frame at time $t$) together with their body orientation. In this paper the detection of the players is given but, alternatively, a detector can be used such as, \eg,~\cite{ren2015faster,cioppa2019arthus,johnson2020sloan}. On the other hand, the orientation of the players in the 2D field is obtained with the method described in~\cite{arbuesOrientation} (for the sake of completeness, details have been given in previous Section \ref{sec:SoA}). From now on, the position and orientation of the players will be considered over a 2D field template. To simplify the notation, the dependence on $t$ of the considered elements will be omitted. 
Let $P$ denote the 2D position in the template field of the player with the ball at time $t$ who is going to execute the pass. Let $\{ R_i, \, i=1,\dots,I\}$ and $\{D_k, \, k=1,\dots,K\}$ denote, respectively, the 2D position in the field of the visible team-mates of $P$, and the current defenders at time $t$, with $I\leq 10, K\leq 11$. The former ones constitute the set of visible receivers of the ball at time $t+\Delta_{t}$, being $\Delta_{t}$ the duration of the pass. 

Let $H_i$ denote the prior or hypothesis that player $P$ is going to pass the ball to receiver $R_i$. The main idea is to define a feasibility measure which is grounded on three elements: (a) the body orientation of every player together with (b) the pressure of the defenders $D_k$, both on $P$ and $R_i$, and (c) the relative position of $R_i$ with respect to $P$.
Then, the most feasible ball pass $\hat{H}$ is computationally selected as the one maximizing  
\begin{equation}\label{eq:maxF}
    \hat{H}=\arg \max_i F(i) ,
\end{equation}
where $F(i)$ is the feasibility of the event pass in $H_i$, which can be defined as
\begin{equation} \label{eq:feas}
    F(i)=F_o(i) F_d(i) F_p(i),
\end{equation}
where $F_o(i)$, $F_d(i)$, and $F_p(i)$ stand for the orientation, defenders and proximity scores, respectively, defined later in this section. Finally, it must be stated that all feasibility measures are obtained right at the moment when the passer $P$ kicks the ball.

\subsection{Orientation} \label{sec:OrComp}
\begin{figure}
    \centering
    \includegraphics[width=0.4\textwidth]{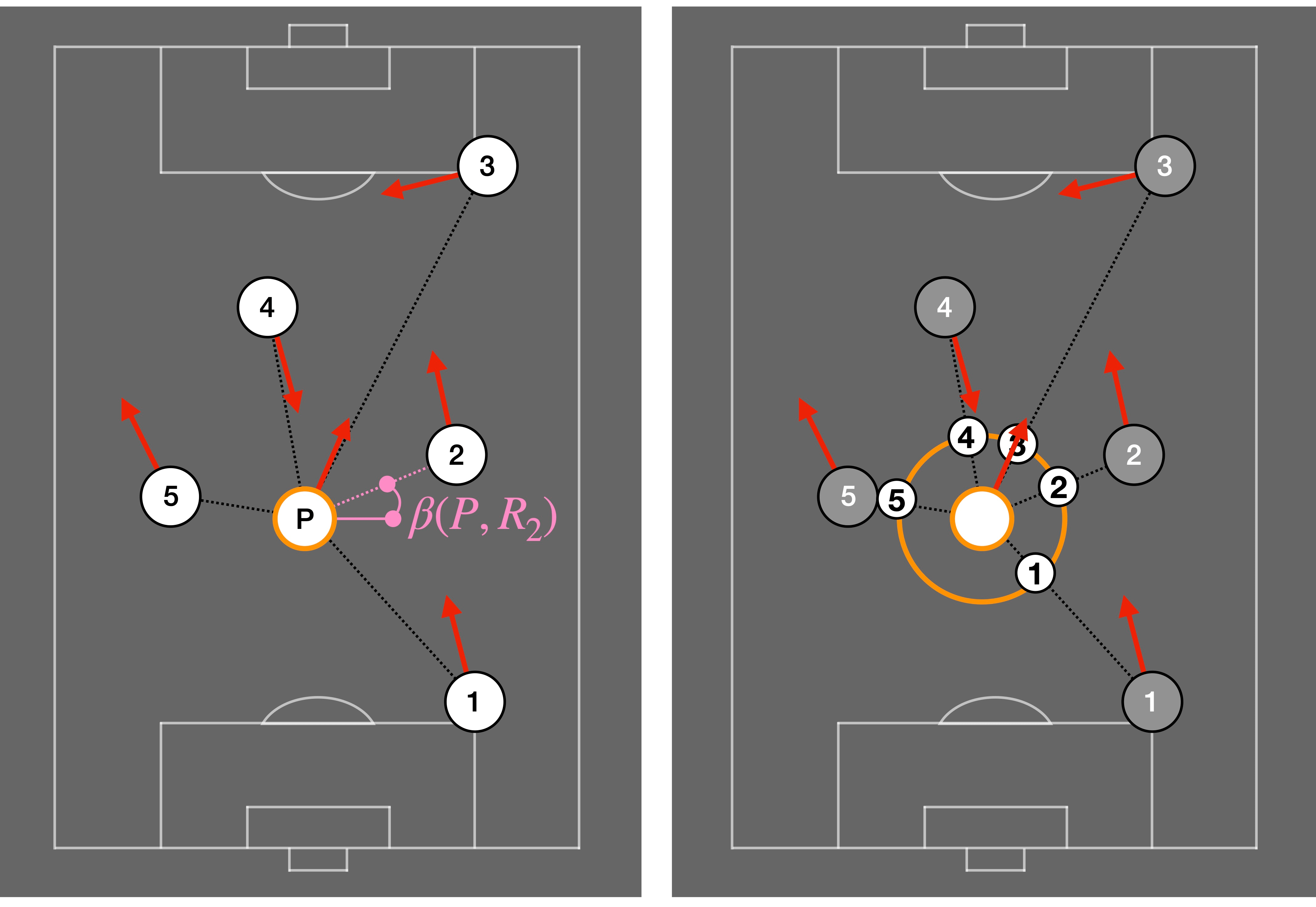}
    \caption{In order not to take pairwise distances into account while computing orientation feasibility, all players are moved towards an equidistant distance (unit circle).}
    \label{fig:Or1}
\end{figure}

One of the aspects that drastically affects the outcome of a pass is the players' body-orientation.  If a player is relatively close to the passer and without being defended, he/she might still not be able to receive the ball properly if he/she is facing away. For a given pass event, the orientation of each player is computed using \cite{arbuesOrientation} in a window of $\pm Q$ frames with respect to the exact pass moment $t$. The median value of these $2Q+1$ observations is considered as the player orientation in the event at time $t$. In practice, a window of 5 frames is used in 25 fps videos. Once obtained this estimation, an orientation-based pass feasibility measure is proposed, which takes into account geometrical quantities and outputs a score of how well a player is oriented in order to receive the ball. 
In order to take only the orientation information into account  (proximity between players will be considered in the 3rd feasibility measure, as seen in Subsection \ref{sec:subDistances}) all potential receivers $R_{i}$ are placed at the same distance with respect to the passer whilst preserving the original angle in the 2D field between the passer $P$ and each receiver $R_i$. Note that this angle is only related to relative position and not to  player body orientation. This step is illustrated in Figure \ref{fig:Or1}.  

\begin{figure}
    \centering
    \includegraphics[width=0.4\textwidth]{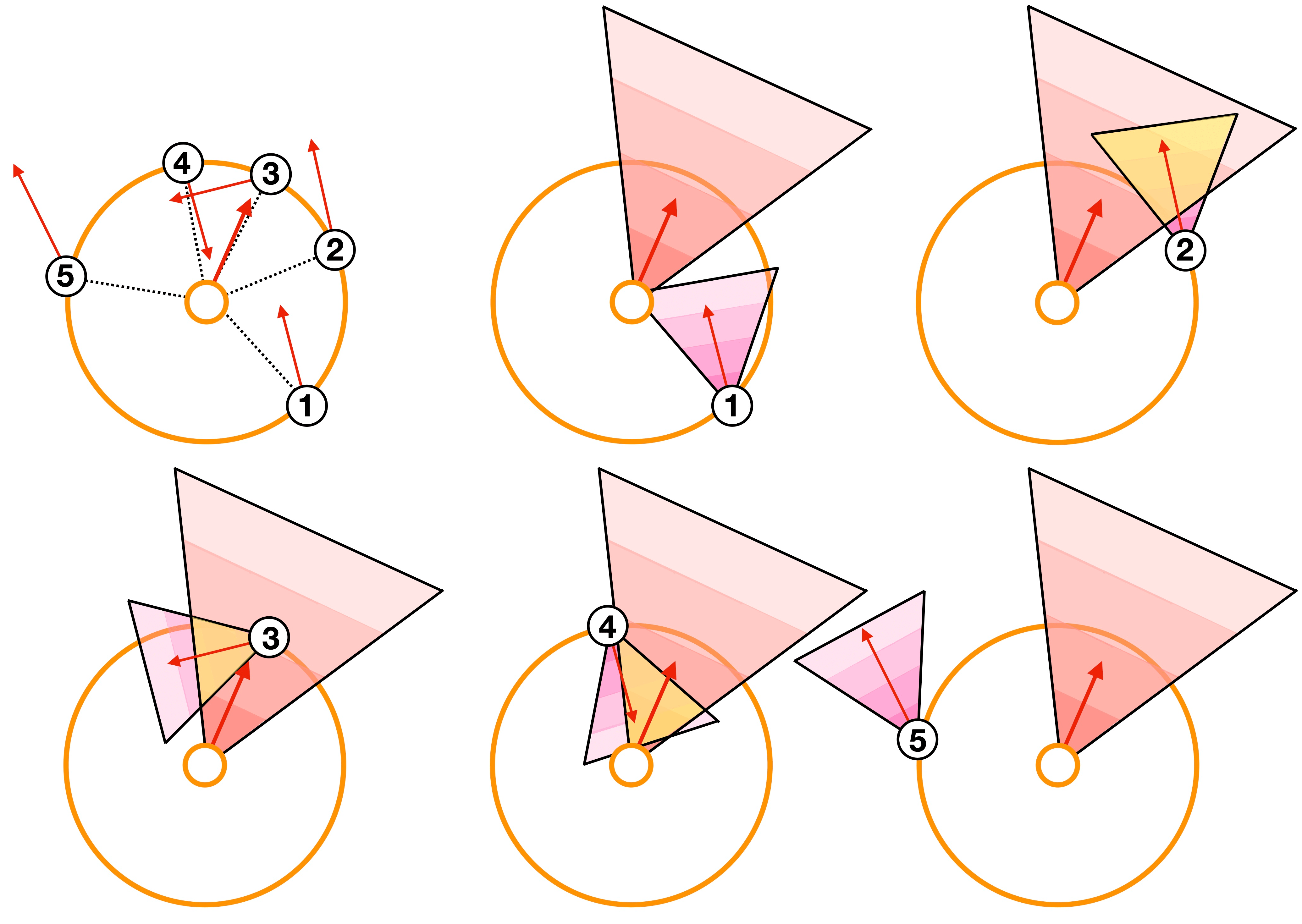}
    \caption{Individual scenarios of intersection given the relocated players of Figure \ref{fig:Or1}. As it can be seen, the top-right player is the  best oriented candidate to receive the ball.}
    \label{fig:Or2}
\end{figure}

Once all potential receivers are placed at an equidistant distance $Z>0$ with respect to the passer, the body orientation of all players, expressed as $\phi(P)$ and $\phi(R_{i})$ for the passer and the receiver $i$ respectively is considered (it corresponds to red vectors in Figures \ref{fig:Or1} and \ref{fig:Or2}). Intuitively, $\phi(P)$ provides an insight of the passer field of view, and by setting a range of $\pm \psi$º with respect to the passer body orientation, an approximate spectrum of the passer field of view is obtained. By setting $\psi$º$>0$ to a fixed value (\textit{i.e.} 30 degrees), an isosceles triangle with the two equal sides of length $2Z$ is defined (see Figure \ref{fig:Or2}). This triangle is denoted by $T_P$ and imposes a limit to the region where the player can pass the ball. The same procedure is repeated for $\phi(R_{i})$, with the triangle $T_{R_i}$ indicating the field of view of the receiver, which shows in which directions he/she can get a pass from; the length of the two equal sides of triangle $T_{R_i}$ is set to $Z$. Figure \ref{fig:Or2} displays some possible scenarios. 
We claim, and numerically verify in Section~\ref{sec:IndPer}, that the weighted area of the intersection of triangles $T_P$ and $T_{R_i}$ gives a measure of how easy it can be for a player to receive a pass in the given configuration: no intersection indicates the inability to get it, whilst partial or total intersection indicates a proper orientation fit. Accordingly, the orientation-based feasibility is defined as
\begin{equation}\label{eq:Fo}
F_{o}(R_{i})=\frac{1}{c}\int_{T_P\cap T_{R_i}} \left(e^{-\text{d}(P,x)}+e^{-\text{d}(R_i,x)}\right) dx
\end{equation}
where $c>0$ is a normalizing constant and $\text{d}(a,b)$ denotes the Euclidean distance between $a$ and $b$ normalized so that the maximum distance in the field is $1$. 

Let us first discuss the weights in \eqref{eq:Fo}. The intrinsic geometry of the triangle has an obvious limitation when it comes to shape intersection: considering the vertex that coincides with the passer position as the triangle beginning, triangles contain a large portion of area in regions placed far from their beginning. Hence, the values inside the computed triangles are weighted according to their relative position with respect to the triangle beginning, fading out in further positions. This effect can be seen as different color opacity in the triangles displayed in Figure \ref{fig:Or2}.
Finally, the reasoning for setting different triangle heights is that, if both passer' and receiver' associated triangles had the same height, players that are located behind a passer who is not looking backwards would intersect notably, despite being a non feasible pass (like in the top-centered example sketch of Figure \ref{fig:Or2}). 

\subsection{Defenders Position}
Apart from considering the visible player  s of the offensive team, the behavior of the defenders, $\{D_{k}\}_k$, is continuously changing the decision-making process. Even if a player is near the passer and properly oriented, the probability of receiving the ball can be really low if he/she is properly guarded; however, it is hard to define how well a player is being defended at a time. Considering only passing events, defenders close to the line that connects the passer with the receiver (passing line) are the ones in a more advantageous position to transform a pass into a turnover. Let us denote by $\beta(P,R_{i})$ the angle in the 2D template field between the passer $P$ and the receiver $R_i$ (see Figure \ref{fig:Or1}), and by $\beta(P,D_{k})$ the one between the passer $P$ and defender $D_{k}$. Using this angle, the proposed defenders-based feasibility will take into account two feasibility scores: (a) the feasibility $\text{F}_{d,P}(R_{i})$ of passing in the direction of $\beta(P,R_{i})$ and (b) the feasibility $\text{F}_{d,R}(R_{i})$ of receiving the ball from  $P$. For the first case, the distance and the angle of all defenders with respect to the passer is computed. 
Therefore, the definition of the feasibility measure $F_{d,P}(R_{i})$ depends on  
the Euclidean distances 
of the closest defenders with respect to the passer:    
\begin{equation}\label{eq:FdP}
\begin{split}
    & F_{d,P}(R_{i}) =\\ 
    & \text{exp}\left(-\frac{1}{J} \sum_{k \in \mathcal{N}_P} w\left(\beta(P,D_{k}),\beta(P,R_{i})\right) (1-\text{d}(P,D_{k}))\right)
\end{split}
\end{equation}
where $\mathcal{N}_P$ denotes the set of the $J$ nearest neighbor defenders from $P$, according to the weighted distance $\text{d}_w$, defined as
\begin{equation}
\text{d}_w (P,D_{k}) =  w(\beta(P,D_{k}),\beta(P,R_{i})) \, \text{d} (P,D_{k})
\end{equation}
where $\text{d} (P,D_{k})$ denotes the normalized Euclidean distance between $P$ and $D_{k}$. Finally, the weights $w$ are defined as
\begin{equation}
 w(\beta(P,D_{k}),\beta(P,R_{i})) = \begin{cases} 0.25 &\mbox{if } \alpha < 22.5 \text{º} \\
0.5 & \mbox{if } 22.5  \text{º} \leq \alpha < 45  \text{º} \\
2 & \mbox{otherwise} \end{cases}
\end{equation}
where $\alpha = |\beta(P,D_{k})-\beta(P,R_{i})|$ (modulus 360º). In practice, we take $J = 3$. 
 
\begin{figure*}
    \centering
    \includegraphics[width=0.85\textwidth]{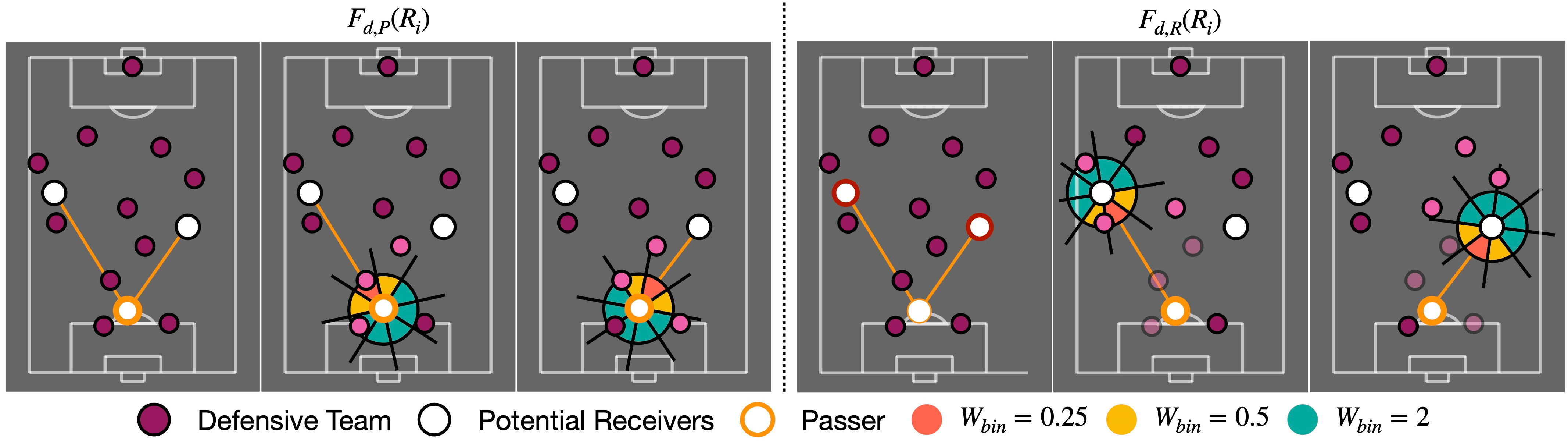}
    \caption{Computation of $F_{d,P}(R_{i})$ and $F_{d,R}(R_{i})$ for two different potential receivers. For both cases, (left) general setup, plus detection of the 3 closest weighted defenders in the scenario of the (middle) left-sided and (right) right-sided player.}
    \label{fig:Def1}
\end{figure*}

Function $w$ is used to model that defenders close to the passing line (and thus with an associated small $\omega$ value) entail a higher risk for that specific pass. This whole procedure can be seen in the left side of Figure \ref{fig:Def1}, where the three closest defenders are highlighted for two hypothetical passes. 

For $\text{F}_{d,R}(R_{i})$, the same procedure is repeated with respect to the receiver; however, in order to have two independent quantities, the $J$ nearest neighbors considered when computing $F_{d}(P)$ are discarded. Hence, 
$\mathcal{N}_{R_i}$ is the set of the $J$ nearest neighbor defenders from $R_i$ (according to $\text{d}_W$) belonging to $\cal{N}_P^C$, \textit{i.e.}, 
the complement of $\mathcal{N}_P$ (that is, the set of the visible defenders at time $t$ that are not in $\mathcal{N}_P$). 
The feasibility to receive the ball from a given angle can be expressed as: 
\begin{equation}\label{eq:FdR}
\begin{split}
& F_{d,R}(R_{i}) = \\
& \text{exp}\left(\!-\frac{1}{J} \!\sum_{k \in \mathcal{N}_{R_i}}\! w\left(\beta(R_{i},D_{k}),\beta(P,R_{i})\right)(1-\text{d}(R_{i},D_{k}))\right)
\end{split}
\end{equation}
The right part of Figure \ref{fig:Def1} shows a graphical example, where the top closest weighted defenders are found with respect to the receiver once discarded the closest defenders found when computing $F_{d,P}(R_{i})$ (Figure \ref{fig:Def1}). 
To conclude, the defenders feasibility is defined as $F_{d}(R_{i}) = F_{d,P}(R_{i})F_{d,R}(R_{i})$, and it is a measure of how likely the event of passing to a particular player is, given the defensive spatial configuration. 

\subsection{Pairwise Distances}\label{sec:subDistances}
Finally, the position in the 2D field affects also the passing options, as players placed closer to the passer have a higher probability of receiving the ball. For this reason, the feasibility of receiving the ball based on pairwise distances or proximity can be defined as inversely proportional to the distance by:
\begin{equation}\label{eq:Fdist}
F_{p}(R_{i}) = \text{exp}\left(-\text{d}(P,R_{i})\right)
\end{equation}

\subsection{Combination} \label{sec:subComb}
Once all three independent feasibility measures are computed, Equation~\eqref{eq:feas} is proposed to combine them. Notice that a low feasibility value in one of the three features (orientation, defenders or distance) indicates that the pass is highly risky, no matter what the other values are.

\section{Results} \label{sec:Res}
The dataset provided by F.C. Barcelona included 11 whole games of their team; not only video footage was provided, but also eventing data. By filtering pass events, 6038 pass events were gathered; these pass events are tagged as well with a binary flag of their outcome, indicating if the receiver was able to control the ball properly (from now on, called successful pass) or not. 
In this Section, several experiments will be detailed with one main goal: to study if proper orientation of soccer players is correlated with successful receptions, thus maximizing the probability of creating a potential goal opportunity. Hence, in order to examine the effect of including the orientation, another baseline pass model will be used for testing, which will only use the output of $F_{p}$ and $F_{d}$; more concretely, $F$ will be compared with $F_{pd}$, defined as:
\begin{equation}
F_{pd}(R_{i}) = F_{p}(R_{i})F_{d}(R_{i}).
\end{equation}
For the whole dataset, in order to measure accuracy, a Top-X metric is obtained by comparing the ground truth receiver of the each pass event with the one indicated by the feasibility scores among all candidates. This metric indicates the number of times (expressed as a percentage) where the current receiver of a given pass is included in the first $X$ candidates according to the feasibility models. In this Section, Top-1 and Top-3 accuracy metrics will be studied under different conditions. Moreover, histograms will be plotted for each scenario. In all cases, the number of bins is 9, as it corresponds to the number of potential receivers of a play; note the goalkeeper has been excluded because it does not appear in the frame domain in many scenarios. The height of each particular bin $B_{n}$ (with $n \leq 10$) represents the number of times that the ground truth receiver has been considered the $n$ best candidate according to the feasibility values (for instance, $B_{1}$ equals the number of times that the actual receiver was considered as the best option). In these Figures, the histograms of successful (blue) and unsuccessful (orange) passes are plotted together.  

\subsection{Orientation Relevance in Pass Feasibility} \label{sec:IndPer}
\begin{figure}[]
    \centering
    \includegraphics[width=0.35\textwidth]{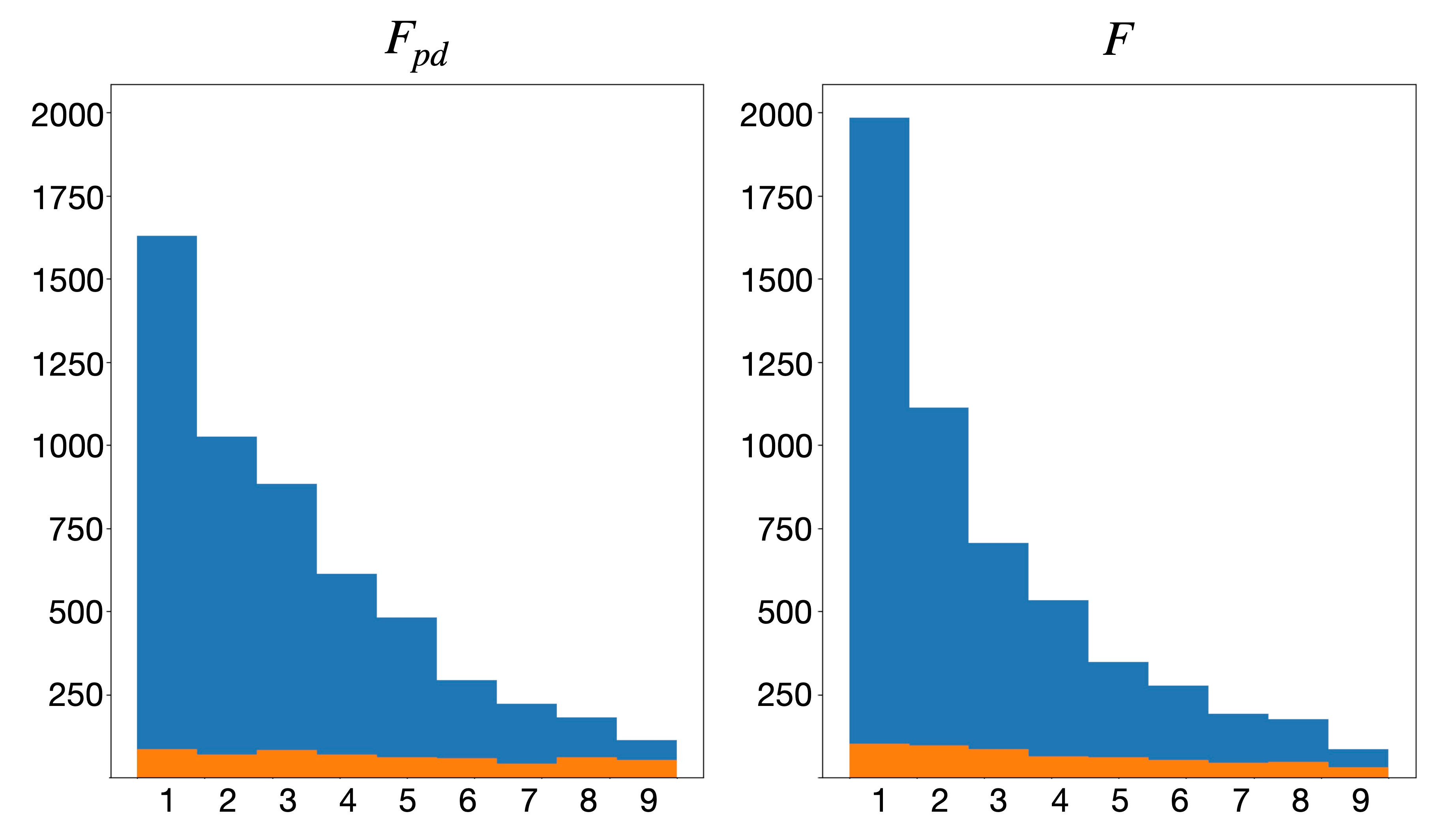}
    \caption{Histogram distribution comparison between $F_{dp}$ and $F$; note that the later includes the computed orientation feasibility. }
    \label{fig:PlotCOMB}
\end{figure}

The importance of orientation in the computation of the proposed feasibility $F$ will be shown by comparing the results of $F$ with the ones obtained with the baseline feasibility $F_{pd}$, which does not include orientation. As it can be seen in Table \ref{tab:tComb}, in both cases the Top-1/3 metric shows that the introduced features in the feasibility computation are directly correlated to the outcome of the play: the difference in Top-1 accuracy between successful and non-successful passes is more than the double, and in Top-3 is more than 0.2. Besides, 
orientation makes a difference by complementing distance and defenders. Apart from boosting the difference between successful and non-successful passes by a margin of 0.04/0.02, $F$ outperforms $F_{pd}$ Top-1 accuracy by 0.07 and Top-3 by 0.05. Visually, this difference can be spotted in the first bins of the histogram displayed in Fig. \ref{fig:PlotCOMB}.

\begin{figure}[]
    \centering
    \includegraphics[width=0.5\textwidth]{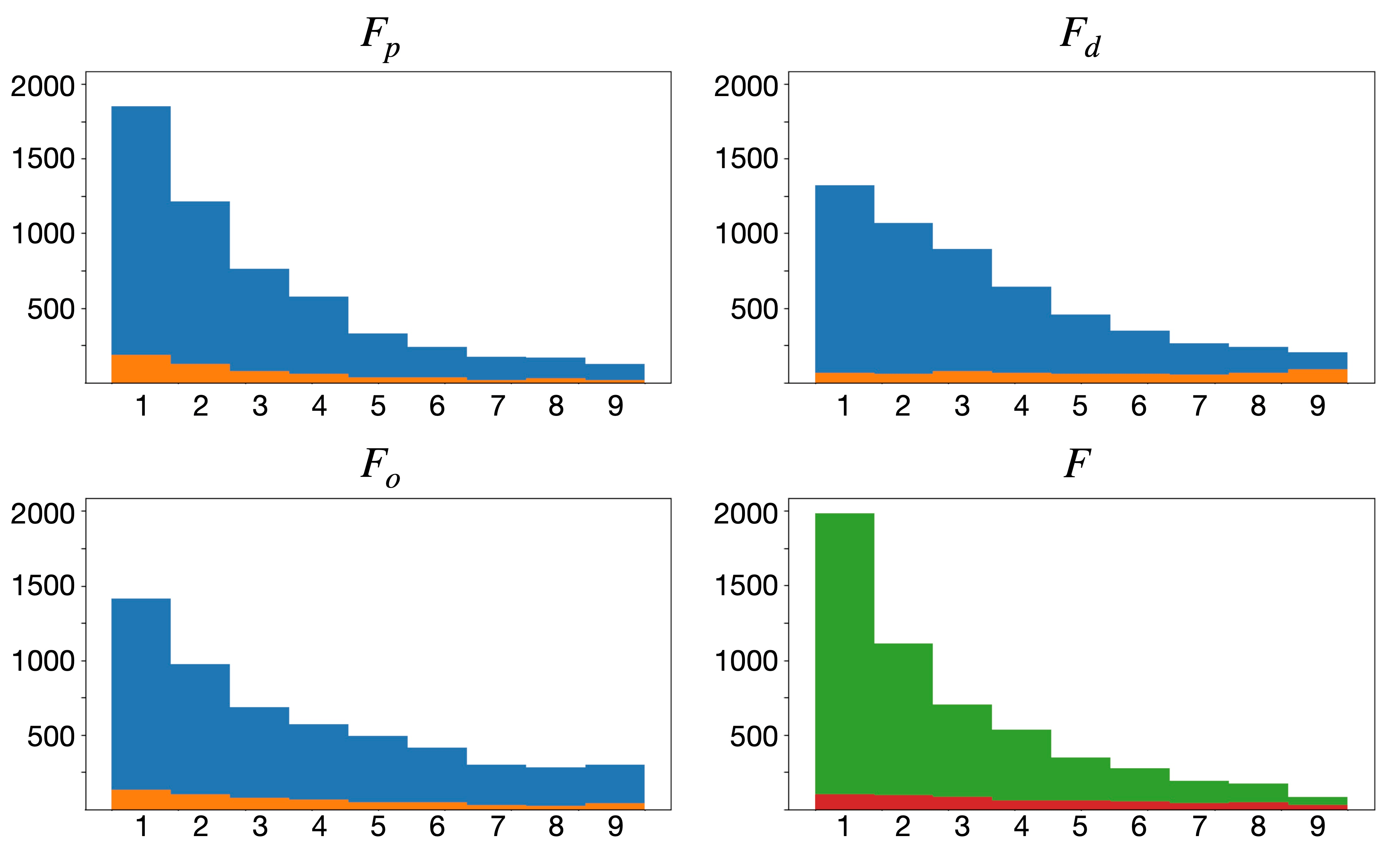}
    \caption{Histogram distribution among potential receivers evaluating individual feasibility components. From left-right, top-bottom: (a) $F_{p}$, (b) $F_{d}$, (c) $F_{o}$ and (d) Combination.}
    \label{fig:PlotIND}
\end{figure}

\begin{table}[]
\begin{center}
\scalebox{0.9}{
\begin{tabular}{|c|c|c|c|c|}
\hline
\textbf{} & \textbf{\begin{tabular}[c]{@{}c@{}}Top-1\\ (Succ.)\end{tabular}} & \textbf{\begin{tabular}[c]{@{}c@{}}Top-1\\ (NSucc.)\end{tabular}} & \textbf{\begin{tabular}[c]{@{}c@{}}Top-3\\ (Succ.)\end{tabular}} & \textbf{\begin{tabular}[c]{@{}c@{}}Top-3\\ (NSucc.)\end{tabular}} \\ \hline
$F_{pd}$     & 0.299                                                            & 0.149                                                             & 0.650                                                            & 0.411                                                             \\ \hline
$F$    & 0.367                                                            & 0.175                                                             & 0.702                                                            & 0.487                                                             \\ \hline
\end{tabular}}
\end{center}
\caption{Top-1/3 accuracy for successful/non-successful passes obtained before ($F_{pd}$) and after ($F$) including orientation as a feasibility measure.}
\label{tab:tComb}
\end{table} 
\noindent\textbf{Decomposed $F_{o}$ - $F_{d}$ - $F_{p}$ Performance.}\\
In order to show how useful the individual estimations are, the performance of the three individual feasibility measures ($F_{p}$, $F_{d}$, and $F_{o}$) is studied together with their  combination. These results are shown in Table \ref{tab:tIndPer} and Figure \ref{fig:PlotIND}. 
For the successful passes, the histogram of all three components share more or less the same shape. However, the top bins of $F_{p}$ have higher values (0.34, 0.70 for Top-1 and Top-3 accuracy respectively); as a result, the bottom bins have low values, which means that it is unlikely to pass the ball to players placed far away with respect to the ball. 
For the unsuccessful passes, $F_{d}$ and $F_{p}$ components seem to be the most and less relevant ones, respectively. This means that passing to a player who is far away does not always imply a turnover, but passing to a well-defended player does (0.14 difference in Top-1 accuracy). Generally, $F_{o}$ resembles $F_{p}$, but the histogram is more distributed (flat shape). 
Combining all three methods (by computing their product) adds some value due to contextualization. For instance, orientation by itself does not take pairwise distances into account: this means that, in particular scenarios, players placed far away in the field might be the best potential candidates in terms of orientation, but as it has been proved, these passes will hardly ever exist. Besides, our proposed feasibility measure $F$ (declared in \eqref{eq:feas}) combines all three components and keeps the high Top-1 and Top-3 metrics of $F_{p}$ whilst preserving the difference between the successful/not-successful passes of $F_{d}$. The bottom-right histogram shows that this goal has been accomplished.

\begin{table}[]
\begin{center}
\scalebox{0.9}{
\begin{tabular}{|c|c|c|c|c|}
\hline
\textbf{} & \textbf{\begin{tabular}[c]{@{}c@{}}Top-1\\ (Succ.)\end{tabular}} & \textbf{\begin{tabular}[c]{@{}c@{}}Top-1\\ (NSucc.)\end{tabular}} & \textbf{\begin{tabular}[c]{@{}c@{}}Top-3\\ (Succ.)\end{tabular}} & \textbf{\begin{tabular}[c]{@{}c@{}}Top-3\\ (NSucc.)\end{tabular}} \\ \hline
$F_{o}$     & 0.260                                                            & 0.232                                                            & 0.566                                                            & 0.546                                                             \\ \hline
$F_{p}$     & 0.340                                                            & 0.320                                                             & 0.704                                                            & 0.665                                                             \\ \hline
$F_{d}$    & 0.243                                                            & 0.107                                                             & 0.604                                                            & 0.336                                                             \\ \hline
\end{tabular}}
\end{center}
\caption{Top-1/3 accuracy for successful/non-successful passes obtained with all three individual feasibility estimations.}
\label{tab:tIndPer}
\end{table}

\subsection{Players' Field Position / Game Phase}
\begin{figure}[]
    \centering
    \includegraphics[width=0.4\textwidth]{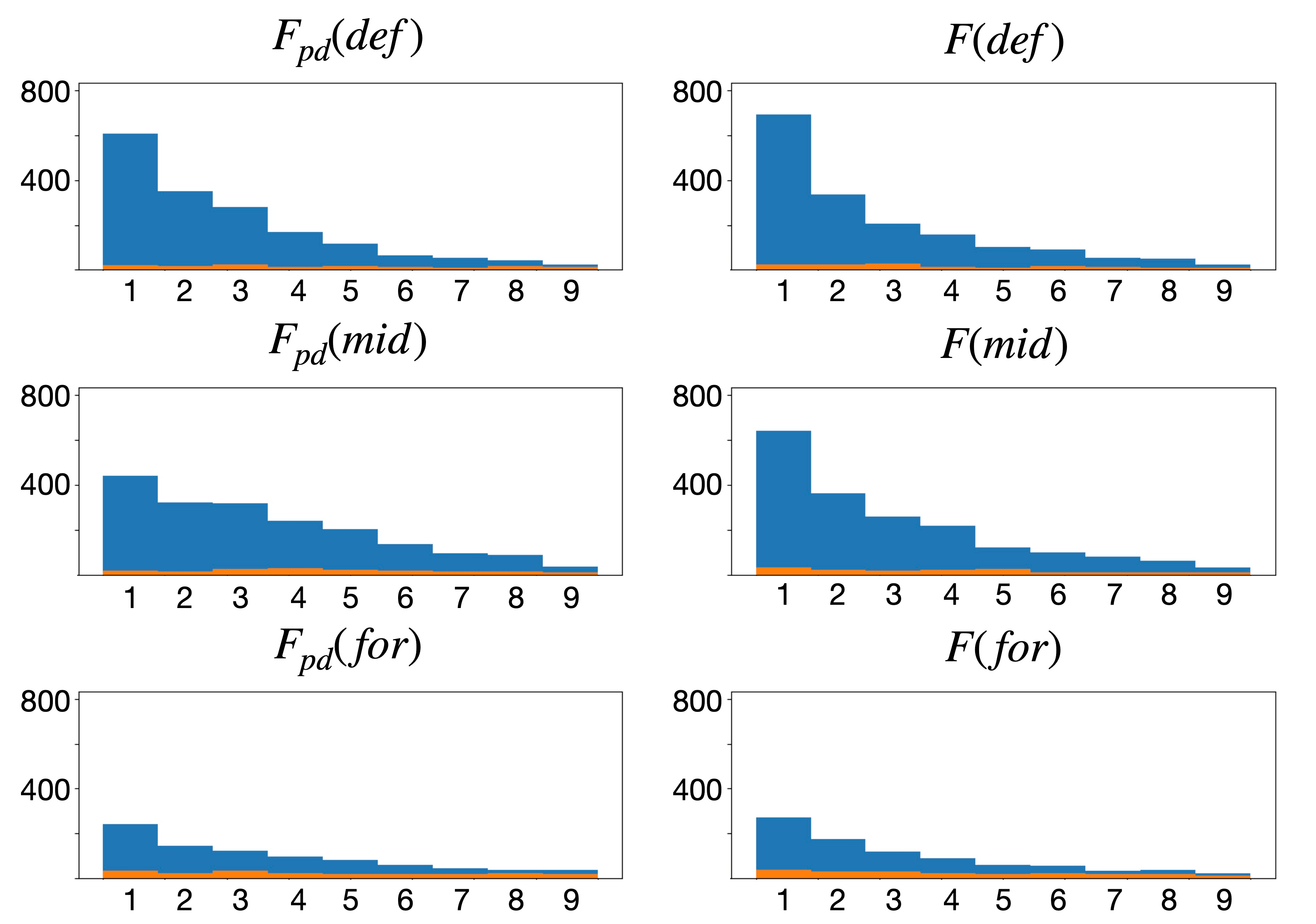}
    \caption{Histogram distribution, obtained with (left) $F_{dp}$ and (right) $F_{dpo}$, for different player positions. From top to bottom: defenders, midfielders, and forwards.}
    \label{fig:PlotPOS}
\end{figure}

Once analyzed the impact of orientation as a feasibility measure, in this Subsection, its effect on different kind of players and game phases is analyzed. 
By classifying them according to the basic field positions (defenders, midfielders and forwards), Figure \ref{fig:PlotPOS} and Table \ref{tab:tType} show the differences, in terms of orientation-based feasibility, among them, which state that midfielders are the ones under bigger $F_{o}$ influence. When introducing orientation in the feasibility measure, both the Top-1 and Top-3 accuracy have a boost of  0.10 while preserving a similar difference in successful-unsuccessful differences (first 3 bins of the midfielders histogram). Defenders are not heavily affected by orientation, mostly because of the many security passes that they perform: in this type of pass (usually between defenders), both players have no opponents surrounding them, and they can freely pass to their closest team-mates without having to be strictly oriented towards them. Forwards are also affected by orientation, but they give and receive less passes; besides, in their domain, passes do not only have a high turnover risk, but also a high potential reward. 

\begin{table}[]
\begin{center}
\scalebox{0.9}{
\begin{tabular}{|c|c|c|c|c|}
\hline
\textbf{}    & \textbf{\begin{tabular}[c]{@{}c@{}}Top-1\\ (Succ.)\end{tabular}} & \textbf{\begin{tabular}[c]{@{}c@{}}Top-1\\ (NSucc.)\end{tabular}} & \textbf{\begin{tabular}[c]{@{}c@{}}Top-3\\ (Succ.)\end{tabular}} & \textbf{\begin{tabular}[c]{@{}c@{}}Top-3\\ (NSucc.)\end{tabular}} \\ \hline
$F_{pd}$ (def.)  & 0.354                                                            & 0.134                                                             & 0.724                                                            & 0.436                                                             \\ \hline
$F$ (def.) & 0.404                                                            & 0.162                                                             & 0.720                                                            & 0.521                                                             \\ \hline
$F_{pd}$ (mid.)  & 0.235                                                            & 0.114                                                             & 0.575                                                            & 0.341                                                             \\ \hline
$F$ (mid.) & 0.341                                                            & 0.196                                                             & 0.673                                                            & 0.456                                                             \\ \hline
$F_{pd}$ (for.)  & 0.278                                                            & 0.158                                                             & 0.589                                                            & 0.426                                                             \\ \hline
$F$ (for.) & 0.315                                                            & 0.178                                                             & 0.653                                                            & 0.459                                                             \\ \hline
\end{tabular}}
\end{center}
\caption{Top-1/3 accuracy for successful/non-successful passes, before/after including orientation, split by player position.}
\label{tab:tType}
\end{table}

\begin{figure}[]
    \centering
    \includegraphics[width=0.3\textwidth]{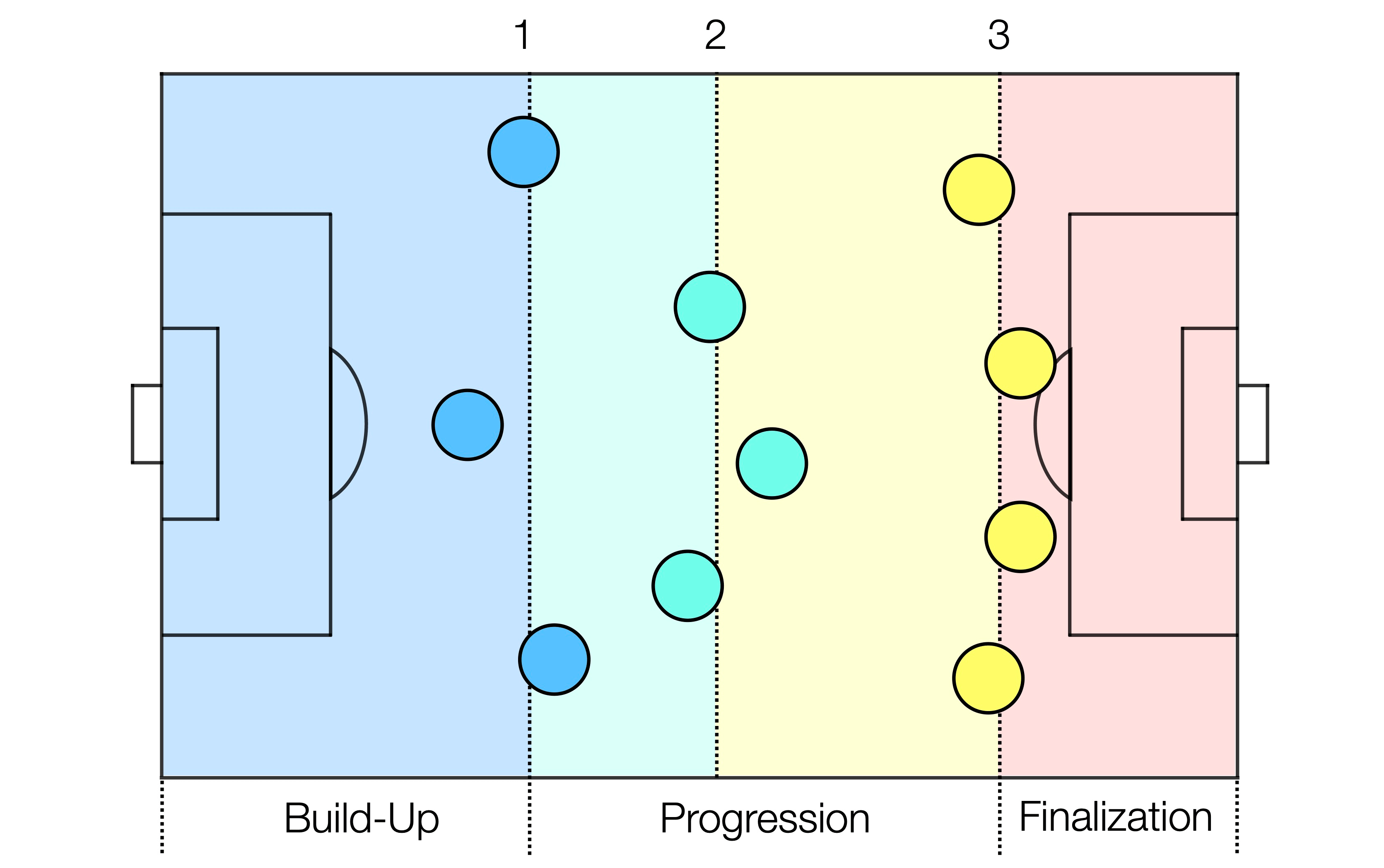}
    \caption{Game phases depend on the position of the passer with respect to the defense spatial configuration.}
    \label{fig:gamePh}
\end{figure}

\begin{figure}[]
    \centering
    \includegraphics[width=0.4\textwidth]{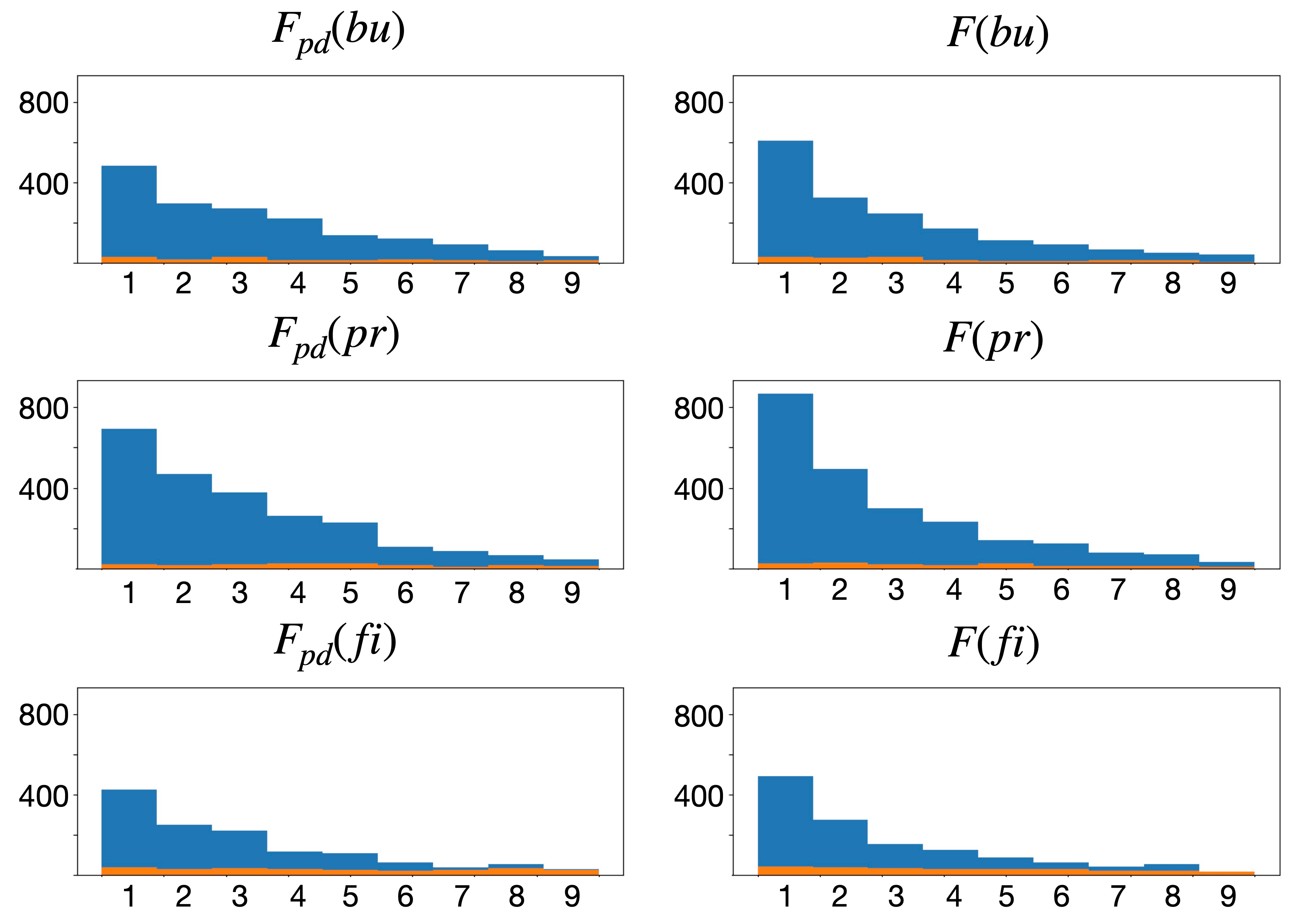}
    \caption{Histogram distribution, obtained with (left) $F_{dp}$ and (right) $F_{dpo}$, for different game phases. From top to bottom: build-up, progression and finalization.}
    \label{fig:PlotPHA}
\end{figure}

In a similar way, passes can be also classified according to the location of the passer in relation to the defensive team spatial configuration, as it is not the same a security pass of a defender than another pass of the same defender but in the offensive side of the court.  In order to introduce this kind of context, different phases of the offensive plays are evaluated individually by clustering the 2D coordinates of the defensive players in the field. Bearing in mind that in a soccer lineup there are mainly 3 rows of horizontally distributed players (both for offense and defense), three phases (displayed in Figure \ref{fig:gamePh}) can be defined: (a) build-up, when the ball is located before the first row of defenders, (b) progression, after the first and before the third row of defenders, and (c) finalization, after the last row of defenders. Results are displayed in Figure \ref{fig:PlotPHA} and Table \ref{tab:tPHA}. Once again, the effect of orientation is vital in the half-court, with a notable difference between successful and non-successful passes in the progression phase (around 0.2 difference in both Top-1 and Top-3, and more than 0.7 Top-3 accuracy). As expected, the build-up and finalization game phases are, respectively, the ones with lower and higher risk, but even in these extreme cases, the inclusion of $F_{o}$ also boosts the pass accuracy metrics.

\begin{table}[]
\begin{center}
\scalebox{0.9}{
\begin{tabular}{|c|c|c|c|c|}
\hline
\textbf{}  & \textbf{\begin{tabular}[c]{@{}c@{}}Top-1\\ (Succ.)\end{tabular}} & \textbf{\begin{tabular}[c]{@{}c@{}}Top-1\\ (NSucc.)\end{tabular}} & \textbf{\begin{tabular}[c]{@{}c@{}}TOP-3\\ (Succ.)\end{tabular}} & \textbf{\begin{tabular}[c]{@{}c@{}}Top-3\\ (NSucc.)\end{tabular}} \\ \hline
$F_{pd}$ (bu.) & 0.282                                                            & 0.143                                                             & 0.610                                                            & 0.382                                                             \\ \hline
$F$ (bu.)   & 0.355                                                            & 0.162                                                             & 0.688                                                            & 0.444                                                             \\ \hline
$F_{pd}$ (pr.) & 0.297                                                            & 0.128                                                             & 0.659                                                            & 0.365                                                             \\ \hline
$F$ (pr.)     & 0.372                                                            & 0.162                                                             & 0.712                                                            & 0.480                                                             \\ \hline
$F_{pd}$ (fi.) & 0.326                                                            & 0.185                                                             & 0.687                                                            & 0.490                                                             \\ \hline
$F$ (fi.)     & 0.376                                                            & 0.203                                                             & 0.710                                                            & 0.534                                                             \\ \hline
\end{tabular}}
\end{center}
\caption{Top-1/3 accuracy for successful/non-successful passes, before/after including orientation, split by player game phase (\textit{bu} - build up, \textit{pr} - progression, and \textit{fi} - finalization).}
\label{tab:tPHA}
\end{table}

\subsection{Combination with Expected Possession Value}
\begin{figure}[]
    \centering
    \includegraphics[width=0.45\textwidth]{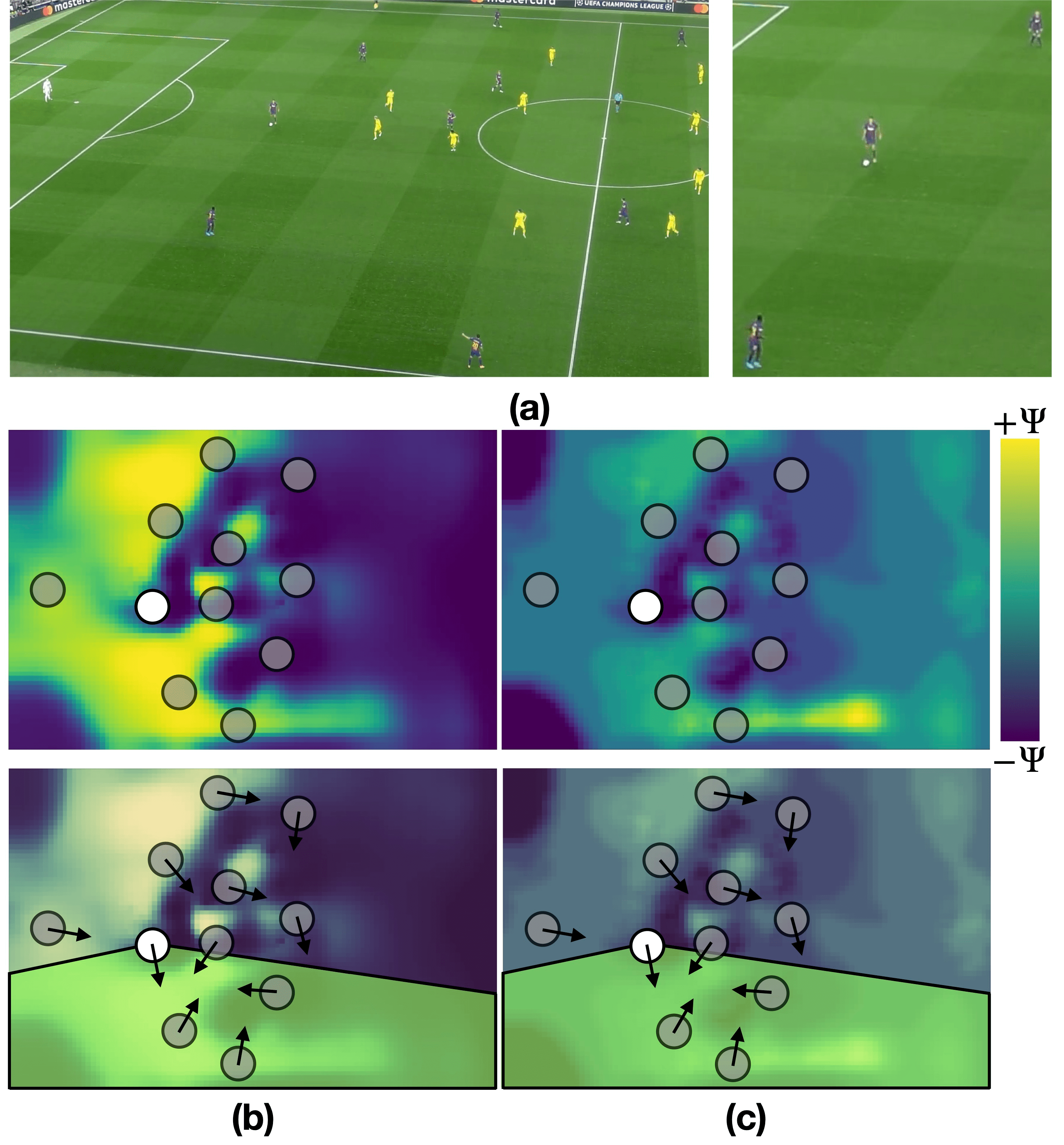}
    \caption{(a) Pass event and zoom in the passer region; (b,c-top) output of the pass probability/EPV models respectively of \cite{fernandez2019decomposing}, typically $\Psi$ equals 0.015, (b,c-bottom) output example made by hand; the combination of the existing models with body orientation would refine the restricting the area of potential receivers.}
    \label{fig:EPVcomb}
\end{figure}

As mentioned in Section \ref{sec:SoA}, EPV is a recently introduced indicator that tries to boost individual/team performance by assigning value to individual actions, using (among others) a pass probability model. However, the EPV model of \cite{fernandez2019decomposing} does not take the body orientation of players into account, thus producing results that, despite being notably accurate, can be refined. An example is shown in Figure \ref{fig:EPVcomb}; for the displayed pass event, the spatial output of the pass probability model (left) and the EPV map (right) can be seen in the middle row. As observed in the original frame, the passer (white circle) is the central mid-fielder, who is directly facing the right-central defender; for this reason, the passer cannot see in his field of view the left-central defender, hence lowering the latter's receiving chances. However, the output of the pass probability model considers the left-central defender as a notable candidate, and EPV does not penalize this pass as a risky one. Nevertheless, by combining our orientation-based feasibility measure $F_{o}$ with the output of the (a) original probability model or the (b) output of the EPV model, maps could be adapted accordingly, thus enhancing potentially good receivers in particular regions as it is displayed in the last row of Figure \ref{fig:EPVcomb}. 

\begin{figure}[]
    \centering
    \includegraphics[width=0.25\textwidth]{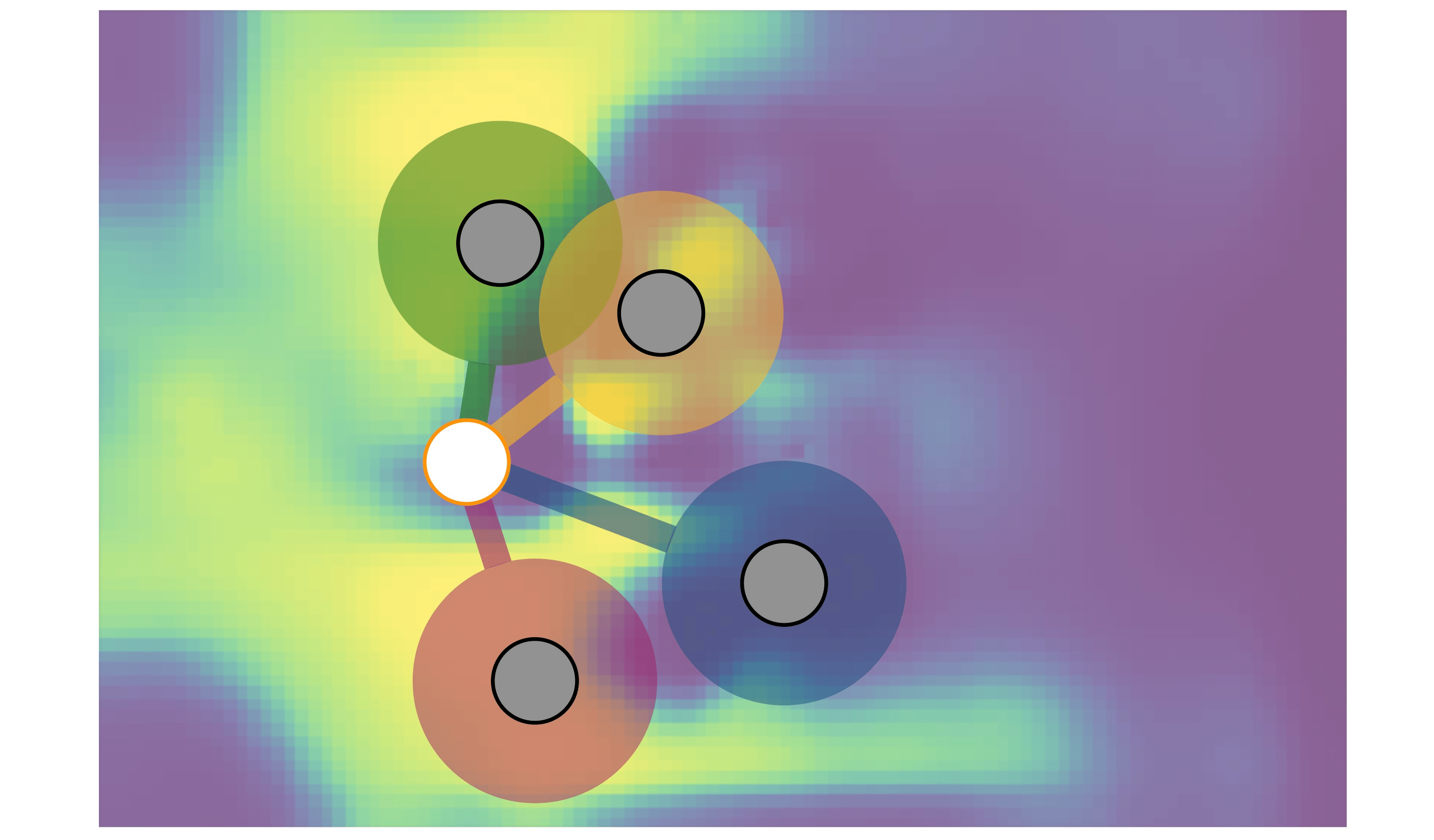}
    \caption{Geometrical approach to assign discretized pass probability/EPV field values to particular potential receivers.}
    \label{fig:EPV2model}
\end{figure}

\begin{figure}[]
    \centering
    \includegraphics[width=0.4\textwidth]{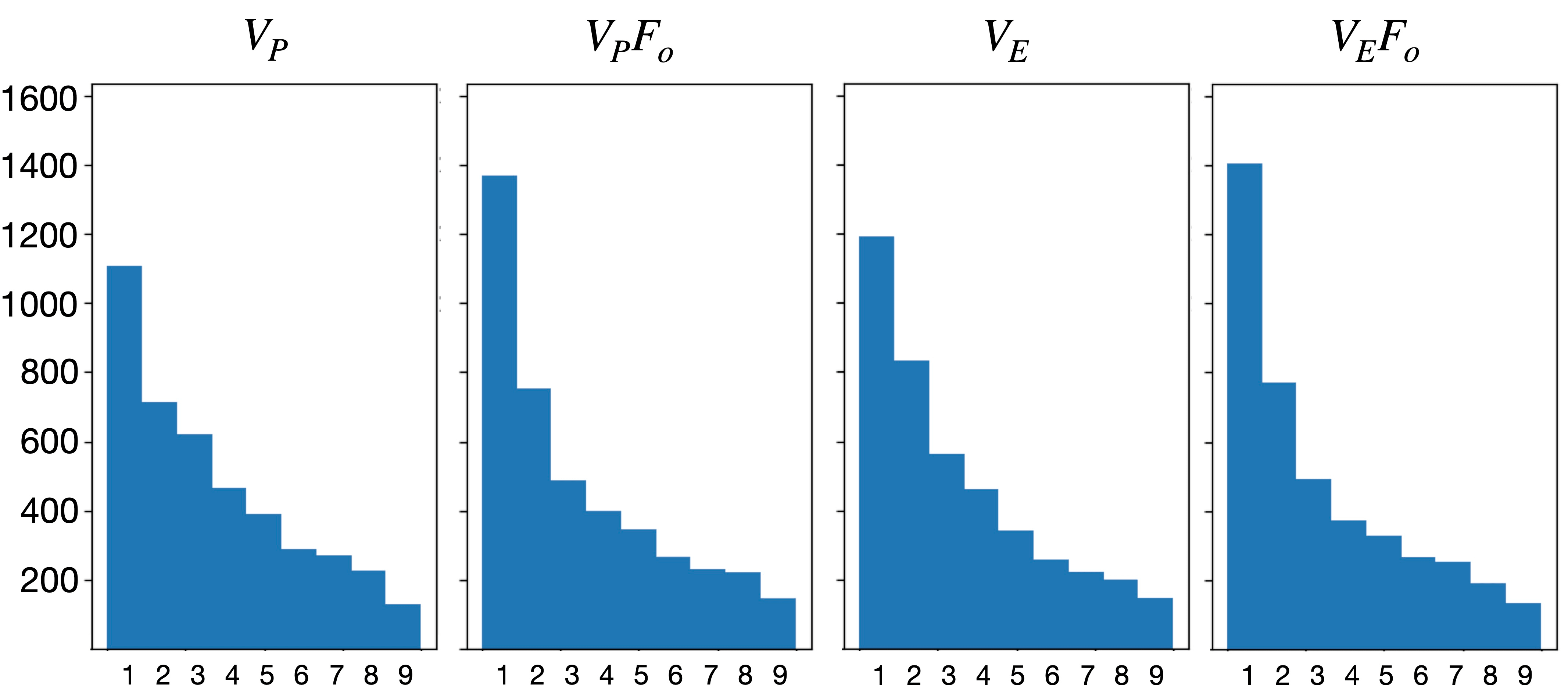}
    \caption{Histogram distribution of $V_{P}$ and $V_{E}$, plus the corresponding addition of $F_{o}$ component.}
    \label{fig:CompPass}
\end{figure}

The main challenge when combining both methods is the dimension miss-alignment: both the pass probability and EPV models extract an output map with a value for each discretized field position (downscaled to $104\times68$), whilst the proposed model defines an individual feasibility value for each of the 10 potential receivers. In order to get a single probability/EPV value for each player in the field, and being $\rho$ the output map (defined by the pixels of the downscaled field), a geometrical solution is provided; its approach is based on the idea that an individual value can be obtained by integrating the probability/EPV values on a meaningful area that extends from the passer to the receiver. In particular, for a given receiver $R_{i}$, first, a disc $Q_i$ of radius $q>0$ is defined around his/her 2D field position, and then, a tubular region $S_i$ of fixed width $s>0$ is defined from  $P$ (starting position) to $R_{i}$ (thus, its length is proportional to the distance between the passer and the potential receiver). The final individual value for receiver $R_{i}$, denoted here as $V(R_{i})$, can be obtained as: 
\begin{equation}\label{eq:SV}
V(R_{i})=\frac{1}{\text{Area}\left(Q_{i}\cup S_{i}\right)} \int_{Q_{i}\cup S_{i}} \rho(x) dx  
\end{equation}
where $\text{Area}\left(Q_{i}\cup S_{i}\right)$ denotes the area of the region $Q_i\cup S_i$. In practice, $q$ and $s$ have been set to $\frac{5}{W_{\rho}}$ and $\frac{2}{W_{\rho}}$, respectively, being $W_{\rho}$ the width of the output map $\rho$ (\textit{i.e.} $104$). Note that Equation \eqref{eq:SV} can be used for both types of maps, being $\rho$ the output of either the pass probability model (from now on $V_{P}$) or the EPV generic model (from now on $V_{E}$). Visually, this whole procedure can be seen in Fig. \ref{fig:EPV2model} for four different receiver candidates. 
For comparison purposes, the individual probabilities $V_{P}$/expected values $V_{E}$ are multiplied by our feasibility orientation estimation $F_{o}$, (Subsection \ref{sec:OrComp}); in this way, the effect of orientation itself can be tested for $V_{P}F_{o}$ and $V_{E}F_{o}$. Note that the other components $F_{p}$ and $F_{d}$ have not been used, as both pass probability and EPV models already include this type of information in its core. Results are displayed in Table \ref{tab:EPVPass} and Fig. \ref{fig:CompPass}. As it can be seen, better accuracy is obtained when taking orientation into account in all scenarios, especially in the top-1 accuracy case, obtaining a boost of almost 0.1 in the output of the current pass probability model. Moreover, orientation also improves the raw performance of $V_{E}$ (0.07 improvement in Top-1 accuracy), especially by solving miss-leading cases in which players are located out of the field of view of the passer. As a conclusion, it has been proved that merging orientation in the SoA implementation of EPV \cite{fernandez2019decomposing} could help getting a more accurate model, which can lead to a better understanding of the decision-making process.  

\begin{table}[]
\begin{center}
\scalebox{0.9}{
\begin{tabular}{|c|c|c|}
\hline
\textbf{} & \textbf{\begin{tabular}[c]{@{}c@{}}Top-1\\ (Succ.)\end{tabular}} & \textbf{\begin{tabular}[c]{@{}c@{}}Top-3\\ (Succ.)\end{tabular}} \\ \hline
$V_{P}$      & 0.243                                                            & 0.567                                                            \\ \hline
$V_{P}$ + $F_{o}$ & 0.332                                                            & 0.612                                                            \\ \hline
$V_{E}$       & 0.266                                                            & 0.606                                                            \\ \hline
$V_{E}$ + $F_{o}$  & 0.337                                                            & 0.637                                                            \\ \hline
\end{tabular}}
\end{center}
\caption{Top-1/3 Accuracy of the EPV models' output, plus their comparison when merging orientation feasibility.}
\label{tab:EPVPass}
\end{table}

\section{Conclusions} \label{sec:Conc}
In this paper, a novel computational model that estimates the feasibility of passes in soccer games has been described. The main contribution of the proposed method is the inclusion of orientation data, estimated directly from video frames using pose-models, into a passing model, which has proved to be a key feature in the decision-making process of players and is strictly correlated to the play outcome. Orientation feasibility is computed with a geometrical approach among offensive players, and it is combined with two other estimations, based on the defenders location with respect to potential receivers, and pairwise distances. Moreover, the combination of the model's output with existing pass probability/EPV models has been studied, obtaining confident results which indicate that SoA methods can be refined by including orientation data. As future work, apart from studying the viability of this type of model in other sports, a passing feasibility discretization of the full-field will be modelled, since players tend to pass not only to the position where the receiver is, but also to large free spaces in front of him/her. Finally, using orientation as a core feature, team action recognition could be applied over the spatial offensive configuration to optimize team tactical strategies. 

\section*{Acknowledgments}
The authors acknowledge partial support by MICINN/FEDER UE project, reference PGC2018-098625-B-I00, H2020-MSCA-RISE-2017 project, ref. 777826 NoMADS, EU H-2020 grants 761544 and 780470 (projects HDR4EU and SAUCE) and F.C. Barcelona's data support.

\end{document}